# Building Large Lexicalized Ontologies from Text: a Use Case in Automatic Indexing of Biotechnology Patents

Claire Nédellec, Wiktoria Golik, Sophie Aubin, Robert Bossy

MIG, INRA, Jouy en Josas
{firstname.lastname}@jouy.inra.fr

**Abstract.** This paper presents a tool, TyDI, and methods experimented in the building of a termino-ontology, i.e. a lexicalized ontology aimed at fine-grained indexation for semantic search applications. TyDI provides facilities for knowledge engineers and domain experts to efficiently collaborate to validate, organize and conceptualize corpus extracted terms. A use case on biotechnology patent search demonstrates TyDI's potential.

## Introduction

The recent development of semantic search engines in specific domains reveals a new need for ontologies that support automatic dense and fine-grained indexing in specific domains. The main limit of the approach is the low availability of adequate indexing resources – linguistic and semantic in specific domains. The work described in this paper takes place in the context of developing semantic search engines in the agronomy domain, for which we use the generic WebAlvis information retrieval system [1]. In this context, we consider the ontology structure as a hierarchy of concepts and the indexing of the text as the annotation of local and independent text phrases by concept labels. The lexical variability and the ambiguity of the terms that denote concepts are processed in the lexical level that bridges the formal conceptual level to the text [2]. The state of the art in section 2 reveals a strong need for termino-ontology acquisition assistants. Section 3 presents TyDI's main functionalities and the methods it allows for the cooperative design of ontologies from texts. Finally, a use case in the biotechnology patents domain shows how TyDI supports the collaboration between a domain expert and a knowledge engineer through a user-friendly interface.

# Context

## Requirements

Domain-specific semantic information retrieval systems require a termino-ontology. A termino-ontology is defined as an ontology comprising a hierarchy of concepts and the most complete set of terminological and lexical manifestations of each concept. The lack of resources is particularly manifest for building scientific and technical information systems. Available terminologies and thesauri generally do not provide formal *is_a* links and most ontologies in the agronomy domain (*e.g.* Gene Ontology [3], the Open Biological and Biomedical Ontologies [4]) are hardly usable for fine-grained indexing because of the domain inadequacy or the lack of lexicalization. Corpus-based extraction of terms is recognized as a rich source of knowledge for termino-ontology design, preserving the explicit links between the text, the lexical and the conceptual level [5]. Term extractors propose term candidates from the text that fulfill linguistic patterns or word co-occurrence criteria. Their application ensures a large coverage of the domain, resulting in lists that can reach several thousands of term candidates. Despite the recent advances in term extraction and ontology learning, such lists still need manual curation and structuring in order to be usable in semantic IR. The efficiency of this work relies on a tool that must allow the selection and the structuring of terms and concepts in a user-friendly way. Ergonomics is particularly crucial for direct use by the expert, so that s/he would focus on domain knowledge acquisition rather than on specific linguistic and formal modeling issues.

## State of the Art

Ontology acquisition from texts has been widely studied in the last decade, resulting in a number of tools that assist the ontology engineer in the various steps of ontology design from term selection to semantic modeling.

Ontology learning frameworks provide integrated pipelines for automatically computing concepts, relations between concepts and inference rules from text analysis, either linguistic or statistical. Several tools have been proposed, among which Text2Onto [], Asium [7], OntoLearn [8] and OntoLT [9].These automatic approaches make useful suggestions that have to be validated and integrated into the knowledge model. Although some of them like as Asium and Text2Onto provide a user validation interface, they do not address the modeling issue with the flexibility of ontology editors.

Closer to our requirements, term validation interfaces and ontology editors assist the manual design of domain specific terminologies and ontologies respectively. The validation of sets of term candidates is often tackled as the simple task of selecting and rejecting words and phrases extracted from a corpus, supported by spreadsheet-like software. It shortly reveals its limits and inappropriateness, especially for term context analysis in the source corpus and cooperative design. The web application TermExtractor [10] stands as an exception as the validation can be compiled from a vote by different users. Unfortunately the validation interface cannot be used as standalone without the associated term extraction tool. At the end of the spectrum, ontology editors like Protégé [11], OboEdit [12] - in the biomedical community- KAON [13] assist ontology creation, edition, browsing, reuse and merging in languages based on formal semantics such as OWL. If they offer user-friendly interfaces for managing concepts and roles, they remain difficult to adopt quickly for simple hierarchy modeling by a domain expert. Finally none of these tools provides facilities for defining concept lexicalization from term candidates extracted from corpus.

The most complete tool with respect to our needs is undoubtedly Terminae [5] that provides an explicit link between the lexical and terminological levels and the conceptual level. It provides a high level of traceability, necessary for the maintenance of the resource. Its limit towards our requirements is the formal distinction between the lexical and conceptual knowledge as materialized by distinct actions and windows in the interface. While this feature makes sense when the ontology design is achieved by knowledge engineers, it turns to act as a hindrance for the domain expert, who has a topic-centered approach of knowledge modeling and needs to transparently handle knowledge from different types at the same time.

## Cooperative Ontology Building from Text

This section details TyDI elementary functions and describes how these functions are combined in methodological steps taking into consideration the cooperative relationship between domain experts and a knowledge engineer.

### TyDI

User cooperation and ontology design traceability relies on a client-server architecture based on the *NetBeans* technology and a PosgreSQL RDBMS for storage. This architecture ensures robustness and flexibility. The input of the TyDI system is a set of term candidates; it supports the YaTeA [14] term extractor format and CSV files. TyDI interface also offers advanced and intuitive navigation and control over the termino-ontology. A TyDI documentation and screenshot are available from [15].

Term candidates are displayed with their linguistic information. The user can specify the columns to be displayed and their order. The content of the term grid is computed by filters specified by the user. The context of each term in the source corpus and its properties can be viewed in specific frames. Distinct frames are dedicated to local structuring and to global modeling. Additional functionalities worth mentioning are the external search facilities (*e.g.* Google, Wikipedia); display of the sub-terms of a given term; exploitation of the term morpho-syntactic variations as computed by the FastR tool [16]; grouped actions via multi-row selection; OWL, OBO and tsv (tab separated values) export.

Term filters include numeric criteria (word count and number of occurrences), used for selecting either short, generally generic terms or long, more specific and generally less frequent terms. The user can select terms with shallow criteria (surface form patterns) or linguistic criteria (surface form, lemma, head, expansion and syntactic category). For instance, one can define filters that displays all terms sharing the same head or all sub-terms syntactically embedded in a given term. These criteria can be combined with Boolean operators. For example, the filter "head term = *tree*" combined sub-terms filter yields the list: *tree*, *mature tree*, *mature avocado*, *mature avocado tree*, *medium-sized mature avocado tree* among which the expert will easily invalidate incomplete term candidates (*mature avocado*) and too specific term candidates (*medium-sized mature avocado tree*).

The filter and sorting functionalities are also useful to build the terminology structure. TyDI facilitates the examination of close terms along various axes. For instance *Head* and *Expansion* columns highlight semantic relations reflected by the term composition: distributional semantic relations (e.g. *insect resistance / bacteria resistance*, *stationary phase / stationary stage*), multidimensional traits (*e.g. plant resistance = resistance of the plant* vs. *insect resistance = resistance to the insects*) and hyperonymy (*e.g. mature tree / mature avocado tree*). These functionalities also help defining synonymy equivalence classes through unconventional term usages, abbreviations, acronyms and rhetoric figures such as metonymy (*e.g. gene activation* vs. *expression activation*) or ellipsis (*e.g. terminator sequence* vs. *terminator*). TyDI allows to qualify specific types of synonymy, *i.e.* acronymy, typographic variation, quasi-synonymy and translation.

A specific frame is used for local modeling of the lexical representation of ontology concepts with terms. Defining a new concept consists in creating a concept label from the term grid or in selecting the label from an existing term synonym class. A dedicated frame provides facilities for structuring the concept hierarchy, for navigating through the ontology and for performing major revisions.

TyDI supports collaborative work by allowing concurrent validations of each term. It maintains in a distinct property the validation status according to each user. The validation status includes a label and a comment where the user can record justifications and questions. The validation status of other users may be visible or hidden depending on the specific application needs; TyDI thus supports open, blind or double-blind validation schemes. Validation statuses also offer filter and sorting criteria, so one can retrieve terms for which a consensus or a controversy exists.

**Strategies for Building Ontologies from Text**

TyDI is independent from methodological strategies of information processing and knowledge modeling: bottom-up, top-down or topic-centered as distinguished in [17]. In fact they are not exclusive and complement each other in the same project as shown in the use case (section 4).

The bottom-up strategy consists in gathering large sets from the whole set of terms extracted from a corpus. Terms are validated in a systematic way by sorting them by shallow properties like the surface form, the number of occurrences or the word count. The most frequent terms, if specific enough, are analyzed in order to scope the main themes. This strategy proves useful for a first approach of the data and for reckoning the data contents.

The top-down strategy aims at structuring the ontology by relating existing subtrees. It starts from ontology overview, and then moves on toward details, by breaking down the model into layers, filling the ontology from high level concepts to more specific concepts. With this approach, modelers have a global insight of the model, allowing them to locate gaps in the model. This strategy is a preliminary step to team work where each participant has specific expertise and is responsible for a particular area of the model.

The topic-centered strategy is more focused: the aim is not to cover a large part of the data, but to focus on local model parts. It starts from a specific term and performs a search in order to assemble a maximum of similar terms within new related classes. The filtering used here is more sophisticated than in the bottom-up strategy, since composition terms analysis and morpho-syntactic variation are extensively used. This strategy is preferred by domain experts who concentrate on a precise issue, usually delaying the strict observation of modeling principles.

The two former strategies prove both useful for modeling at the leaf level (bottom-up approach) and at more abstract levels (top-down approach), while the latter (topic-centered) is convenient for locally populating ontology labels and synonym classes.

**A Cooperative Process**

TyDI is intended for both domain experts and knowledge engineers who collaborate relying on complementary skills. Thus TyDI is designed for concurrent cooperative work. Our main hypothesis is that the domain expert shall play an active role in knowledge modeling through a direct use of the tool. We believe that a strong involvement of the domain expert is both efficient and gratifying. We also assume this involvement is feasible with the help of appropriate methodologies of cooperation between the expert and a knowledge engineer, and the use of an appropriate tool as an interface between them. We discuss here the respective contribution of the expert and the engineer and how TyDI can act as the medium tool. These general principles will be illustrated in the use case in section .

In this context, the design of a termino-ontology requires that both the expert and the knowledge engineer acquire some knowledge of the objective and constraints of the other party. The knowledge engineer (KE) enforces the target application and processing tools constraints. S/he also trains the expert to use TyDI functionalities at different acquisition phases. By learning how to use the tool, the expert is allowed to grasp the practical consequences of these constraints, as well as some of the underlying principles of terminology and formal modeling. S/he validates domain terms using his/her background knowledge and context facilities of TyDI while the KE relies on linguistic clues, composition related functionalities and document context to check the well-formedness of validated terms.

During the construction of synonym equivalence classes, the expert proposes preferred terms and synonyms, while the KE discusses the preference according to the term frequency and linguistic constraints such as length. S/he also checks that synonymy and hyperonymy relations are strict enough and do not denote weak relatedness with respect to the application requirements. S/he may suggest additional synonyms by analyzing similar terms and term context in the corpus.

The specific cases of ellipses, metonymies and metaphoric usage are debated in order to distinguish hyperonyms, synonyms and non-related variants. The debate is ideally collegial and requires references and inspiration from external resources of the application domain. During this interaction the KE acquires the domain basic notions empowering him to be more autonomous so the expert can concentrate on more complex issues. Finally the KE role is also to help expert in choosing the best strategy for ontology design (see section 3.2). For instance, when the expert goes too deep in specific topics and looses the view of the global direction, the KE may suggest alternative strategies such as top-down structuring of the existing knowledge chunks.

## Use Case: Ontology Building from Biotechnology Patents

We present here a practical experiment of collaborative ontology design supported by TyDI. The target application is the semantic search of patents in the biomass exploitation domain for the VegA project [18]. The patent belongs to the ECLA patent class A01H: *New plants or processes for obtaining them; plant reproduction by tissue culture techniques,* referred to as *New Plant* domain in the following. That work was set up to meet the actual needs of intellectual property engineers and biology experts in technology watch and evaluation of freedom to operate. The IR application is now operational and publicly available [19]. The patent collection consists of the 21,039 OEB patent documents in English as available from esp@cenet web service. The *title*, *abstract* and *claims* sections were used for the term extraction and then indexed by the termino-ontology. YaTeA software extracted 65,529 terms imported into TyDI. A plant biology expert and a knowledge engineer collaborated to the *New Plant* termino-ontology building. They interacted remotely by using TyDI interface to visualize the results of their actions in real time, and by email and phone for planning. The design consisted of four phases: (1) shallow term filtering, (2) topic-based exploration of the terms, (3) term validation, classification and local modeling, and (4) global modeling and concept formalization. According to section 3.2, the two first phases belong to the *bottom-up* strategy, the third part is representative of the *topic-centered* strategy and the fourth one is an example of the *top-down* strategy.

### Shallow, Non Semantic List Filtering

Given that the term candidates were automatically extracted, the list contained incorrect and irrelevant terms. The first stage consisted in cleaning the terms list by shallow filtering and was mostly performed by the KE. She attempted both to delete most incorrect forms and to group inflexions and derivations into synonyms classes. Since the list contained many morphological variants (e.g. *public controversies*, *public controversy*, *Public controversy*) that were missed by the lemmatization step, the engineer wrote explicit rules to merge them automatically, the merged terms were still visible on the TyDI interface for traceability purposes. After merging, the number of terms decreased by about 11% (7,403 terms). The most frequent incorrect forms, were identified by sorting the terms of the surface form column by alphabetical order that highlights badly segmented terms starting with special characters ( *%,[,"* ). Other numerous irrelevant terms were identified thanks to the surface form filter: rhetorical terms and terms relative to the genre of documents (*in addition, claim, main objectives)*, terms containing irrelevant adjectives or demonstratives (*previous study, this disease*), over-general and ambiguous terms (*intensity*, *product*, *concentration*).

A lot of hapax, often resulting from incorrect segmentation or POS tagging were found out by sorting terms by their occurrence frequency and tagged as invalid.

### Topic-based Terms Exploration

The goal here was to identify important domain themes. To filter the terms, the expert and KE used shallow clues such as frequency and length of terms. Frequency sort highlights representative terms of recurrent themes in the corpus (ex. *DNA sequence:* 542 occ*., transgenic plant:* 664 occ.). They also sorted terms by number of words, giving priority to short (monolexical) terms to identify general items (*e.g. polymer, breeding, phenotype, disease*) or, on the contrary, giving priority to longer multilexical terms to identify domain specific items (*e.g. herbicide resistant acetohydroxyacid synthase*). In parallel, the interaction with the KE became more fluent as the expert familiarized with TyDI, learnt its main functionalities as well as the terminology and formal ontology fields. Concurrently, the KE discovered the main biological notions.

### Terms Validation, Terms Classification and Local Modeling

In this phase, the expert selected a seed term as representative of one relevant theme (*e.g. maize cell*), then he browses through terms by following different types of semantic roles in order to cover thoroughly the theme of the elected theme. The user combined different filtering criteria at the surface level (*e.g.* regular expression match) or at the linguistic level (same head, same expansion). This kind of unbounded navigation allowed the expert to discover alternative semantic axes (*maize cell* vs. *maize culture*), or semantically close elided terms (*maize plant cell*).

Since, a lot of terms turned out to be semantically close, the expert benefited from the filter display to gather them within classes. To check the meaning of an ambiguous term, he displayed its context and queried search engines from TyDI interface. We observed that when checking for valid terms candidates, the expert is naturally impelled to group and structure the terms within semantic classes and subclasses, and thus to conceptualize them into a local model in a continuous schema. Even though this blurs the formal frontier between the lexical and the conceptual levels, for efficiency reasons the distinction should not be enforced at the operational level in the user interface. The expert handled both: terms linked by synonymy relationship (*virus resistance, viral resistance* and *resistance to virus*) choosing a class representative (*virus resistance*) and classes linked by *is_a* relationship

(*geminivirus resistance, potyvirus resistance is_a virus resistance)*.

Thanks to the multi-window view, the expert was able to quickly populate existing synonyms and hyperonyms classes and to create new ones by looking for other terms displayed within the grid. Moreover the expert distinguished different synonymy sub-cases corresponding to linguistic variation types: typographic variants (*indole butyric acid* and *indolebutyric acid*) and acronyms (*IBA*). Other synonyms resulted from previous machine processing (*e.g.* OCR error: *Brassica oleracea*, *Brassica oieracea*).

### Global Modeling and Concepts Formalization

At this stage, the respective roles of expert and knowledge engineer became well-defined: conceptualization and modeling of term classes belonged mostly to the domain expert, whereas the KE made sure generality relations were strict and consistent with formal knowledge modeling rules, and also helped the expert resolve consistency problems. The term classes produced during the previous stage were used as a basis for ontology design. Each class representative was reevaluated as denoting an ontology concept, while outclass terms were systematically considered as potential candidates for denoting new concepts. Supported by the engineer, the expert massively gathered synonyms that were spread across the terminology (e.g. *Brassica iuncea* and *Brassica iuncea plant* with their synonyms). As already observed in previous modeling experiences, the restriction to only *is-a* relations sometimes led the expert to gather related concepts in informal concept bags instead of formalizing them into different hierarchies, *e.g. resistance to infection*, *to disease*, *to pathogen*. The formalization of such bags depends on the application requirement. In fact, in the A01H domain the distinction between the three types of resistance is considered as irrelevant for semantic document searching. Alternatively, he gathered concepts related to different issues because of the similarity of their labels, *e.g. solar energy* and *metabolizable energy* denote respectively the source of the energy and the capability of the plant to metabolize it. The KE systematically reviewed the topics investigated by the expert in order to overcome these problems.

During that stage, the expert proceeded by creating several general concepts, then by populating the ontology with more specific concepts and enriching more in depth issues in which the expert is particularly versed. In order to correct the bias of the single expert view and the potential lack of representativeness of the training corpus, an expert of the biomass domain handed to the domain expert an additional list of topics representative of the documents. He also relied on relevant external information resources, *i.e. Mesh* and *Agrovoc* for structuring the highest levels of the ontology, choosing relevant concept labels.

### Ontology Building Results

The design of the *New Plant* termino-ontology took 6 weeks, which is very short compared to our comparable previous semantic search application developments with Yatea using a spreadsheet and Protégé instead of TyDI. 21,960 (37%) terms were validated and 10,603 (18%) terms were deleted. 5,967 (10%) terms belong to 2680 synonym classes which are linked to the same number of hierarchical concepts.

The development of an instance of WebAlvis allowed us to evaluate the quality of a sample of the termino-ontology through an IR application. The evaluation results are available from [15]. This evaluation showed the high quality of the resource with regard to formal and lexical criteria, thus validating our methodology.

## Conclusion

The improvement of semantic information retrieval systems, especially in highly specific domains, depends on the quality of the knowledge representing resources developed. Specific domain knowledge acquisition from corpus is a work in progress and requires appropriate methodologies and enhanced tools.

We specified that fine-grained document analysis applications (Information Retrieval, Information Extraction) require formal ontologies with rich lexicalization. This kind of resource is hardly available and its acquisition is very time-consuming.

We proposed a methodology for building lexicalized ontologies where the domain experts are strongly involved, to the point they also contribute to the modeling of the resource. We demonstrated that this involvement can be efficient with the help of an appropriate methodology and a specialized tool to help knowledge engineers and domain experts to work collaboratively. We presented the TyDI software through the use case of a patent retrieval application for the biotechnology domain. We are currently improving TyDI based on feedback collected during this experiment. Immediate future developments will focus on handling polysemy and integrating available ontologies and terminologies. Long terms improvements include ontology learning features (distributional semantics, Hearst-like patterns) and better collaboration support (history management, locks, conflict resolution functions).

We are involved in several projects in different domains that will provide more use cases, giving the opportunity to improve both TyDI and our methodology.

**Acknowledgement.** The authors acknowledge of TyDI development by Frédéric Papazian. They also thank Bernard Teyssendier, the domain expert for his contribution to the *New Plant* ontology. This work is funded by Quaero program.